\documentclass[10pt,twocolumn,letterpaper]{article}

\usepackage[pagenumbers]{cvpr} %

\usepackage{tikz}
\usepackage{graphicx}
\usetikzlibrary{spy, calc} %

\usepackage{colortbl}
\usepackage{caption}
\usepackage{subcaption}
\usepackage{textcomp}
\usepackage{gensymb}
\usepackage{xspace}
\usepackage{enumitem}
\usepackage{tabularx}
\usepackage{hhline}
\usepackage{booktabs}
\usepackage{wrapfig}
\usepackage{soul}
\usepackage{multirow}
\usepackage{ctable}
\usepackage{pifont}
\usepackage{xurl}
\usepackage{pgfplots}

\newcommand{\pin}{\textbf{P}_\text{in}}
\newcommand{\pnv}{\textbf{P}_\text{nv}}

\renewcommand{\paragraph}[1]{\vspace{.5em}\noindent\textbf{#1}}

\definecolor{cvprblue}{rgb}{0.21,0.49,0.74}
\usepackage[pagebackref,breaklinks,colorlinks,allcolors=cvprblue]{hyperref}

\def\paperID{****} %
\def\confName{CVPR}
\def\confYear{2026}

\title{Stepper: Stepwise Immersive Scene Generation with Multiview Panoramas}

\author{
Felix Wimbauer$^{1,3,4,\dagger}$\and
Fabian Manhardt$^{1}$\and
Michael Oechsle$^{1}$\and
Nikolai Kalischek$^{1}$\and
Christian Rupprecht$^{2}$\and
Daniel Cremers$^{3,4}$\and
Federico Tombari$^{1,3,4}$\and
$^{1}$Google\quad\quad
$^{2}$University of Oxford\quad\quad
$^{3}$MCML\quad\quad
$^{4}$Technical University of Munich
\and
{\tt\small felix.wimbauer@tum.de\quad fabianmanhardt@google.com}
}       

\begin{document}

\twocolumn[{%
\renewcommand\twocolumn[1][]{#1}%
\maketitle
\vspace{-2.5em}
\begin{center}
\includegraphics[width=.95\textwidth]{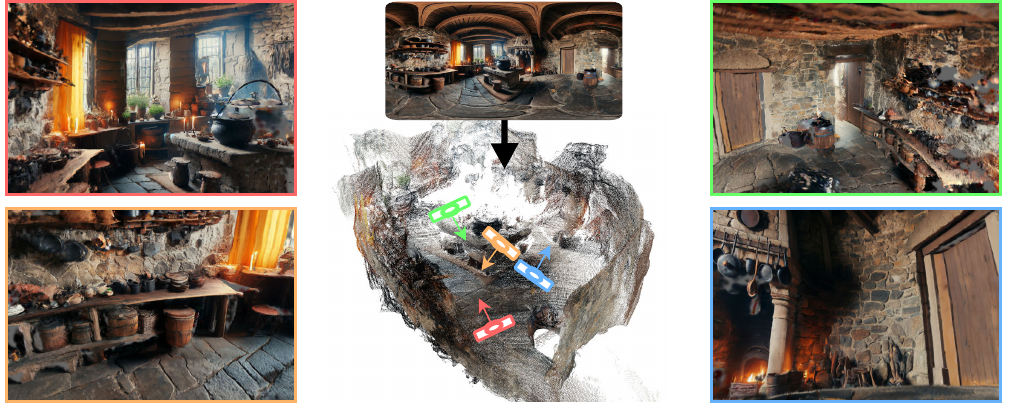}
\vspace{-.3cm}
\captionof{figure}{\textbf{Stepper} sets a new state-of-the-art quality level of generated explorable 3D scenes. Its core innovation is a novel cubemap-based multi-view panorama diffusion model that ensures high-resolution scene synthesis while facilitating step-wise, coherent scene expansion and high-quality scene reconstruction.
Please check out our project page at: \href{https://fwmb.github.io/stepper}{fwmb.github.io/stepper}
}
\label{fig:teaser}
\end{center}
}]
\maketitle
\begin{abstract}
The synthesis of immersive 3D scenes from text is rapidly maturing, driven by novel video generative models and feed-forward 3D reconstruction, with vast potential in AR/VR and world modeling. 
While panoramic images have proven effective for scene initialization, existing approaches suffer from a trade-off between visual fidelity and explorability: autoregressive expansion suffers from context drift, while panoramic video generation is limited to low resolution. 
We present {\em Stepper}, a unified framework for text-driven immersive 3D scene synthesis that circumvents these limitations via stepwise panoramic scene expansion. 
Stepper leverages a novel multi-view 360° diffusion model that enables consistent, high-resolution expansion, coupled with a geometry reconstruction pipeline that enforces geometric coherence. 
Trained on a new large-scale, multi-view panorama dataset, Stepper achieves state-of-the-art fidelity and structural consistency, outperforming prior approaches, thereby setting a new standard for immersive scene generation.

\end{abstract}
    
\section{Introduction}
\label{sec:intro}

\let\thefootnote\relax\footnotetext{${}^\dagger$Work done during Felix' internship at Google.}
The synthesis of immersive 3D scenes from text or images has rapidly evolved from a novel challenge to a central task in computer vision, mirroring the unprecedented success of generative image and video models.~\cite{yang2025matrix, yang2025layerpano3d, schneider2025worldexplorer} This task serves as an essential stepping stone towards general world models, but has immediate applications in spatial computing, particularly for Mixed Reality and next-generation mapping applications. Crucially, in these settings, synthesized environments must satisfy strict perceptual criteria: high-fidelity rendering, visual consistency and unrestricted navigation within the synthesized environment.

To address this challenge, recent work has largely focused on two distinct paradigms. The first adopts an iterative strategy that leverages generative image or video models to autoregressively hallucinate and fuse novel views into an expanding scene representation~\cite{hollein2023text2room, cai2023diffdreamer, chung2023luciddreamer,fridman2023scenescape,schult24controlroom3d}. While theoretically enabling large-scale exploration, this approach is susceptible to subtle inconsistencies and context drift, often resulting in accumulating geometric errors and degraded visual fidelity. Alternatively, a second line of work~\cite{zhou2024holodreamer, schwarz2025recipe,schneider2025worldexplorer,yang2025matrix} targets lifting 360° panoramas directly into 3D space. Although these methods deliver superior visual quality near the projection center, they fundamentally struggle with occluded regions; rendering viewpoints far from the origin inevitably introduces artifacts such as blurring and stretched primitives.

To bridge this gap, we propose a method that enables the generation of high-quality 3D scenes, as seen in \cref{fig:teaser} with support for large-baseline navigation, achieved through three primary contributions. First, we introduce a multi-view panorama diffusion model. Leveraging an initial canonical view from~\cite{kalischek2025cubediff}, this model enables us to essentially "step" into the scene. Crucially, by processing full panoramic contexts rather than limited field-of-view perspective images, our approach minimizes the accumulation of geometric and semantic inconsistencies, \ie context drift. Simultaneously, it circumvents the resolution bottlenecks of panoramic video generation, delivering the high-definition imagery that ensures superior immersion.

Second, we introduce a reconstruction framework that enforces geometric consistency across multiple panoramic views. To avoid distortions and undesired artifacts inherent to conventional monocular depth estimators on spherical data, we decompose our generated multi-view panoramas into perspective views and process them with a robust feed-forward SfM model~\cite{keetha2025mapanything} to recover a dense point cloud. Subsequently, we optimize a 3D Gaussian Splatting~\cite{kerbl3Dgaussians} representation for real-time exploration.

Third, we release a large-scale synthetic dataset to overcome the severe scarcity of multi-view panoramic data. Existing public collections suffer from limited scale, low resolution, and a lack of multi-view observations required to learn scene exploration. We extend the procedural generation framework Infinigen~\cite{infinigen2023infinite, infinigen2024indoors} to render high-quality, multi-view panoramic trajectories across a diverse set of indoor and outdoor environments.  Comprising approximately 230\,000 samples at $4096\times2048$ resolution across 5,000 scenes, this dataset provides the geometric priors necessary for strong generalization. Additionally, we also curate a small test set of 3D scenes from Infinigen and the web, allowing us to benchmark our model against existing baselines. To summarize, our contributions are:
\begin{enumerate}
    \item \emph{A multi-view, high-resolution panorama diffusion model} for iterative scene expansion, 
    \item \emph{A robust reconstruction framework} that synthesizes generated multi-view panoramas into a consistent, explorable representation.
    \item \emph{A large-scale, multi-view panorama dataset} with an accompanying benchmark set for improved training and rigorous evaluation of existing world models.
\end{enumerate}

\section{Related Work}
\label{sec:relatedwork}

In this section we introduce all relevant related work. To this end, we first start by discussing 2D generative models, before diving into the literature for 3D reconstruction and synthesis from images.

\subsection{2D Generative Models}

\paragraph{Image Generation.} With the introduction of diffusion models~\cite{sohl2015deep, ho2020denoising, song2020denoising}, the field of image generation has recently taken a huge leap forward. Facilitated by advances in latent space modelling~\cite{rombach2022high} and classifier-free guidance~\cite{ho2022classifier}, modern models~\cite{labs2025flux1kontextflowmatching, podell2024sdxl, esser2024sd3} are capable of synthesizing photo-realistic images at high resolution from merely text prompts with reasonable hardware requirements.
Methods like LoRA finetuning~\cite{hu2022lora}, ControlNets~\cite{zhang2023adding} and Adapters~\cite{mou2024t2i} allow users to introduce explicit constraints, enabling tasks ranging from inpainting~\cite{corneanu2024latentpaint, lu2025pinco} to layout-guided synthesis~\cite{zheng2023layoutdiffusion}.

\paragraph{Controllable Video Generation.} Building upon image foundations, diffusion models have recently demonstrated remarkable success in video synthesis~\cite{wan2025wan, hong2022cogvideo, yang2024cogvideox, genmo2024mochi, kong2024hunyuanvideo, openai2024sora, openai2025sora2, wiedemer2025veo3}. To leverage these models for scene exploration, research has focused on disentangling camera movement from content generation. By injecting explicit trajectory and intrinsic control into the denoising process, recent approaches~\cite{wang2024motionctrl, he2024cameractrl, yu2024viewcrafter, mark2025trajectorycrafter, zhang2025matrix} can be used for controllable exploration of unobserved parts of a virtual scene.

\paragraph{Panorama Generation.} Parallel to video generation, a distinct line of work focuses on 360$^\circ$ panorama synthesis~\cite{feng2023diffusion360,wang2023panodiff, wu2024panodiffusion, gao2024opa, lu2024autoregressive}, which offers a compact yet comprehensive representation of scene context and can serve as a strong 3D scene initialization. However, most existing methods rely on generating equirectangular panoramas, which introduces significant polar distortions and restricts resolution. To circumvent these limitations, CubeDiff~\cite{kalischek2025cubediff} proposes repurposing multi-view diffusion models to jointly synthesize the six faces of a cubemap. We identify this cubemap-based paradigm as a robust foundation for 3D scene exploration and adopt it as the backbone for our proposed multi-view panorama generation model.

\subsection{3D Representations from 2D images}

Obtaining a 3D scene representation from a set of images and videos is a long-standing problem in computer vision. 

\paragraph{Monocular Depth Estimation.} Historically, 3D reconstruction relied on estimating explicit geometry from single views~\cite{eigen2014depth, laina2016deeper}. This field has recently matured into the era of foundation models; pioneered by MiDaS~\cite{Ranftl2022midas}, current approaches demonstrate that training on massive, diverse datasets unlocks strong zero-shot generalization~\cite{yang2024depthanythingV1}. Modern depth predictors~\cite{ke2024repurposing, bhat2023zoedepth, yang2024depthanythingV2} are able to infer highly-detailed geometry from unconstrained images, including metrically accurate depth maps~\cite{piccinelli2025unik3d, piccinelli2024unidepth, piccinelli2025unidepthv2, yin2023metric3d, hu2024metric3dv2, wang2025moge}.

\paragraph{Joint Pose and Pointmap Estimation.} While depth maps provide local geometry, reconstructing a coherent scene requires aligning multiple views in 3D space. Recent advances have precipitated a paradigm shift from traditional Structure-from-Motion pipelines to end-to-end foundation models. Methods like DUSt3R~\cite{wang2024dust3r} and its successors propose to simultaneously regress camera poses and dense 3D pointmaps directly from image sets. Current state-of-the-art methods~\cite{wang2025vggt, keetha2025mapanything} achieve robust reconstruction across large numbers of input images, leveraging the rich priors distilled from extensive, densely annotated training datasets.

\paragraph{Novel View Synthesis.} To translate these geometric representations into immersive exploration, the field has largely adopted neural rendering techniques. Neural Radiance Fields (NeRFs)~\cite{mildenhall2021nerf, barron2021mip, barron2023zip} established the standard for photorealistic view synthesis, while 3D Gaussian Splatting (3DGS)~\cite{kerbl3Dgaussians, ye2025gsplat, niemeyer2025radsplat} has recently emerged as a dominant alternative, offering real-time rendering speeds with high visual fidelity. In this work, we bridge these paradigms by employing MapAnything~\cite{keetha2025mapanything} to lift our generated multi-view panoramas into a consistent geometric scaffold, which is subsequently optimized into a 3DGS representation for real-time exploration.

\subsection{3D Scene Synthesis}

\paragraph{Perspective-based Synthesis.} Synthesizing coherent 3D environments from sparse inputs is a fundamentally ill-posed problem. Nevertheless, recent methods have made progress by leveraging advances in generative models and 3D reconstruction. Approaches like DiffDreamer~\cite{cai2023diffdreamer}, Text2Room~\cite{hollein2023text2room}, and others~\cite{fridman2023scenescape, chung2023luciddreamer, schult24controlroom3d} adopt an iterative paradigm, using depth estimation to lift images in 3D and image diffusion to autoregressively fill missing regions, with extensions to interactive scene generation~\cite{yu2024wonderjourney, yu2025wonderworld, wulff2025dream}. These methods inherently rely on partial observations, leading to severe semantic drift and geometric inconsistencies when moving far from the origin. Alternative video-based methods~\cite{liang2025wonderland} attempt to generate exploration paths directly but often struggle to maintain multi-view consistency without explicit geometric constraints.

\paragraph{Panoramic Scene Synthesis.} To overcome these limitations, recent research has shifted toward panoramic generation. Methods like HoloDreamer~\cite{zhou2024holodreamer} and RfG3D~\cite{schwarz2025recipe} lift a single generated panorama into 3D but degrade rapidly during translation due to the lack of disoccluded geometry. WorldExplorer~\cite{schneider2025worldexplorer} and Matrix-3D~\cite{yang2025matrix} address this by driving exploration with panoramic video models, with the latter achieving improved stability via a feed-forward 3DGS reconstruction model. However, the computational cost of video synthesis imposes severe resolution bottlenecks, limiting output fidelity. Inspired by the multi-view success of CAT3D~\cite{gao2024cat3d} and CubeDiff~\cite{kalischek2025cubediff}, we propose to overcome these trade-offs by treating scene expansion as a multi-view cubemap generation problem. By iteratively synthesizing high-definition cubemaps rather than low-resolution panorama videos, our method minimizes drift while maintaining the photorealistic quality required for immersive exploration.

\section{Method}
\label{sec:method}

\begin{figure*}[t!]
\centering
\includegraphics[trim=1.2cm 0 0 0, clip, width=\textwidth]{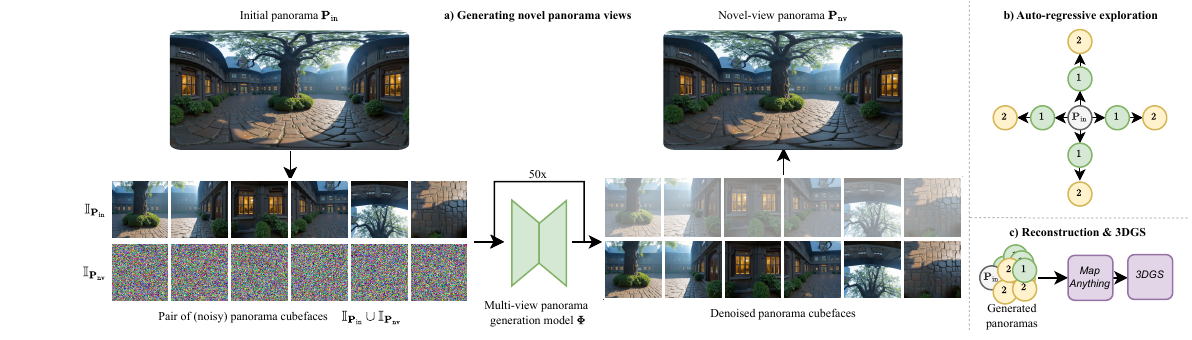}
\vspace{-.75cm}
\caption{\textbf{Method overview.} a) Our model generates a new panoramic image from a previously unobserved viewpoint based on a given input panorama. To ensure high quality, we utilize a pre-trained diffusion model with expanded multi-view attention that is instrumental for jointly denoising the high-resolution cubefaces of the newly generated novel-view panorama.
b) Our ability to generate novel-view panoramas enables auto-regressive scene generation in all directions of the scene yielding a set of high quality, consistent panoramas that effectively complete the representation of the 3D scene. c) The generated panoramas are processed with the feed-forward reconstruction model, \textit{i.e.} MapAnything. The output pointcloud serves as the initialization of a custom 3D Gaussian Splatting reconstruction enabling high quality novel view synthesis of the generated 3D scene. 
}
\label{fig:main_method}
\end{figure*}

In the following, we first introduce the required preliminaries before describing our multi-view panorama generation model for step-by-step 3D scene exploration.
Finally, we present a specialized pipeline, visualized in \cref{fig:main_method}, to obtain an immersive 3D scene from multiple exploration steps.

\subsection{Preliminaries}
\label{ssec:prelim}

Let $\mathbf{P} \in [0, 1]^{H\times W\times 3}$ be a panoramic image in the equirectangular format, which covers the full sphere with $360^\circ\times180^\circ$.
A 3D point $\textbf{x}\in \mathbb{R}^3$ can be mapped to the image plane using the panoramic projection function
\begin{equation}
    \pi_{\text{pano}}(\textbf{x})=
\begin{pmatrix}
    \frac{W}{360} \cdot \arctan\left(\textbf{x}_z, \textbf{x}_x\right) \\
    \frac{H}{180} \cdot \arctan\left(\textbf{x}_y, \sqrt{\textbf{x}_x^2 + \textbf{x}_z^2}\right)
\end{pmatrix}.
\end{equation}
A perspective image $\textbf{I}\in[0, 1]^{h\times w\times 3}$ with camera rotation $\textbf{R}\in \operatorname{SO(3)}$ and focal length $f$ can be obtained from $\textbf{P}$ through resampling (and vice versa)
\begin{equation}
    \textbf{I}_p(\textbf{P}, \textbf{R}, K) = \textbf{P}\left[\pi_{\text{pano}}\left(\textbf{R} K^{-1}\begin{pmatrix}p_x \\ p_y \\ 1\end{pmatrix}\right)\right],
\end{equation}
with $K$ denoting the camera intrinsics given the focal length $f$, and $\textbf{I}_p$ being the resulting perspective image at pixel $p$. 

\subsection{Multi-view Panorama Generation}
\label{ssec:mvpano_gen}

Most 3D scene synthesis methods start from an input image or panorama and then rely on off-the-shelf 2D image or video priors to fill in unobserved areas of the scene.
However, as these models are usually conditioned on regular perspective images with a small field of view, they often struggle to grasp the entirety of the scene.
This results in context drift and 3D inconsistencies.
We argue that operating directly on panoramic images presents a promising alternative.
First, panoramas, when projected to the right representation, are still 2D and fairly similar to perspective images, and can thus strongly benefit from pretrained 2D image and video models.
Second, panoramic images always capture a significant portion of the scene context, ensuring more coherence and reducing drift.

To this end, we propose a multi-view panorama generation model $\Phi_d: \pin \mapsto \pnv$, which synthesizes a novel panoramic view by virtually moving a distance $d$ forward into the scene from the input panorama $\pin$.
Such a model can thus be naturally utilized for full 3D scene exploration. By rotating the input panorama $\pin$ by an angle $\alpha$, which can be achieved by horizontally rolling the equirectangular image by $\alpha \cdot \frac{W}{360}$ pixels, we can adjust the movement direction. Through auto-regressive invocation, we can then take multiple steps and hence walk along longer paths.

\paragraph{Multi-cubemap representation.}
As the availability of multi-view panoramic image pairs is limited, it is not feasible to train such a model from scratch and still achieve good generalization capabilities.
Therefore, we adopt a pretrained image diffusion model for conditional panorama generation.
Drawing inspiration from \cite{kalischek2025cubediff}, we represent panoramic images as cubemaps by sampling the six faces of a cube $\mathbb{I}_\textbf{P} = \left\{\textbf{I}(\textbf{P}, \textbf{R}_{\text{k}}, K) \mid k \in \left\{\text{Front}, \text{Left}, \ldots, \text{Bottom}\right\} \right\}$.
Note that we set $f$ in $K$ such that each face has a FOV of $90$ degrees. Hence, a given equirectangular panorama $\textbf{P}$ can be fully described by these perspective images and vice-versa.
A pair of panoramas is then the set $S \in [0, 1]^{12 \times H\times W\times 3}$ of all twelve cubefaces with $S = \mathbb{I}_{\pin} \cup \mathbb{I}_{\pnv}$.
Thanks to this representation, we can now feed our panoramas to regular image diffusion models by simply setting the batch size $t=12$, without suffering from a domain gap due to distortions.

\paragraph{Model architecture.}\label{sec:model}
Similarly to \cite{gao2024cat3d}, we base our model on an LDM with a latent space of $128\times128\times8$, which is pretrained on a large-scale image dataset. 
In particular, the LDM follows an architecture similar to Stable Diffusion~\cite{rombach2022high} with multiple convolutional and self-attention layers. 
Inspired by~\cite{kalischek2025cubediff}, we modify the architecture to simultaneously generate multiple images, \textit{i.e.} the different cubefaces $\mathbb{I}_\textbf{P}$.
To ensure cross-view and cross-panorama consistency, we inflate~\cite{shi2023mvdream} the
deeper self-attention layers of the LDM in order to enable the tokens of every cubeface to attend to the tokens of all other faces of its own as well as the other cubemap. 
In practice, the self-attention token sequence length is simply extended for those layers from $(bt)\times (hw)\times l$ to $b\times(thw)\times l$, where $b$ is the number of panoramas in the batch, $hw$ denotes the spatial dimensions, and $t=12$, as noted above, denotes all cube faces in $S$.
Note that we also concatenate a positional encoding $p$ and mask $m$ to each pixel in order to encode the pixel location and whether it needs to be generated.
We compute UV coordinates on the unit cube and mark each pixel by its panorama of origin with
\begin{equation}
    p = \pi_{\text{pano}}(x) \quad \text{and} \quad
    m_p = 
    \begin{cases}
        -1 & \text{if } p \in \pin \\
        1 & \text{otherwise}
    \end{cases},
\end{equation}
where $x$ denotes the 3D coordinates of the point on the cube face.
Note that we do not need to condition the LDM on $d$ as we empirically found that a fixed stepping length works best for scene expansion.
Finally, we finetune the model using a standard diffusion loss on ground truth panorama pairs converted to cubefaces.

\subsection{3D Gaussian Splats from Panoramas.}
\label{ssec:rec_from_pano}
While the described model already enables us to freely generate 3D-coherent panoramas at different viewpoints, it does not yet allow real-time exploration of the scene.
To this end, we additionally distill our multi-view panoramas into a 3D Gaussian Splatting (3DGS)~\cite{kerbl3Dgaussians} representation for real-time novel view synthesis.

\paragraph{Pointcloud from feed-forward model.}
Recently, several works have shown that a strong initialization prior, in the form of a 3D pointcloud, can be helpful for high-quality results and training stability~\cite{niemeyer2025radsplat, kotovenko2025edgseliminatingdensificationefficient}.
However, as aligning individual depth maps from a given monocular depth prior can be prone to errors, we instead apply a state-of-the-art feed-forward reconstruction model MapAnything~\cite{keetha2025mapanything} on perspective views that we extract from all of our generated panoramas.
Nevertheless, the application of MapAnything to cubemap faces is not trivial.
First, the lack of overlap with other views can lead to unsatisfying results, which is particularly prominent in the up- and downward facing views.
To better mimic MapAnything's training data distribution, we design a different viewing pattern for reconstruction: we rotate 45 degrees up and down from the horizontal cubefaces to always ensure sufficient overlap among views.
Note that we show resulting 3D pointclouds for different input patterns to MapAnything in the supplementary material.
Second, the resulting point maps can easily become very large with a huge number of redundant points, leaving a large memory footprint and slowing down rendering. Hence, in an effort to remove these redundant points, we build the final pointcloud in an iterative fashion.
Using the pointcloud renderer from PyTorch3D~\cite{ravi2020pytorch3d}, we check if newly introduced points by a panorama are already visible in any of the previous panoramas.
We then only add the previously unobserved points to the final pointcloud.

\paragraph{3DGS optimization.}
For photometric reconstruction, we use projected views of generated panoramas, consisting of the six cubefaces and eight additional perspective views.
Further, we initialize the 3DGS representations with the accurate pointcloud from the feed-forward model and apply a simplified optimization strategy of MCMC-GS~\cite{kheradmand20243d}. Due to the under-constrained nature of our setup and the dense initialization, we reduce the complexity of the optimization problem by assuming fixed Gaussian positions and only a color value per Gaussian as the appearance representation.
For details of the 3DGS optimization, we refer to the supplementary materials.

\subsection{Step-wise Scene Exploration}
\label{ssec:scene_exp}

Our framework relies on a starting panorama $\textbf{P}_\text{init}$ to initialize the scene for 3D exploration.
If no panorama is available, we rely on the state-of-the-art method CubeDiff~\cite{kalischek2025cubediff} to generate a high-resolution panorama image $\textbf{P}_\text{init}$ given a text prompt and/or a reference image.
From the initial panorama, we take $n$ auto-regressive steps using our model $\Phi$ into four directions to obtain $1 + 4n$ panoramic views, covering significant portions of the scene, which were initially unobserved in $\textbf{P}_\text{init}$.
Finally, by lifting these panoramas to 3D Gaussians, as described above, we enable full real-time 3D exploration of the generated scene.

\section{Experiments}
\label{sec:experiments}

In the following, we first describe the design of our training and test datasets, before qualitatively and quantitatively evaluating our approach against state-of-the-art baselines.

\subsection{Data \& Setup}
\label{ssec:data}

\begin{figure}[t!]
\centering
\includegraphics[width=\linewidth]{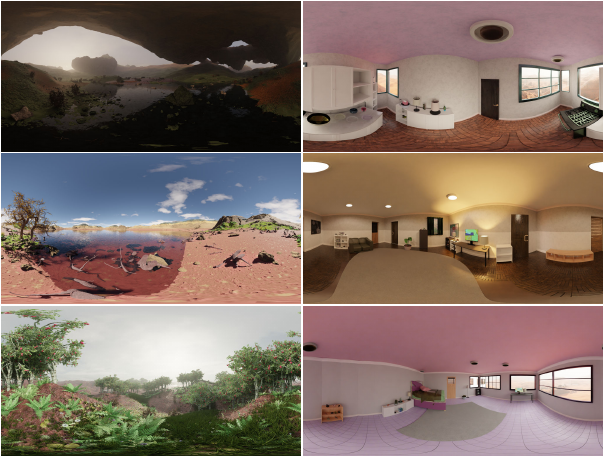}
\vspace{-.6cm}
\caption{\textbf{Dataset Samples.} The dataset generated with Infinigen consists of a diverse set of high quality synthetic panoramas of indoor and outdoor scenes. For every panorama, we rendered a pair from a novel viewpoint enabling the training of the multi-view panorama generation model. All panoramas are aligned to the horizontal line.}
\label{fig:infinigen_data}
\vspace{-.4cm}
\end{figure}

\begin{figure*}[t!]
\centering
\includegraphics[width=\textwidth]{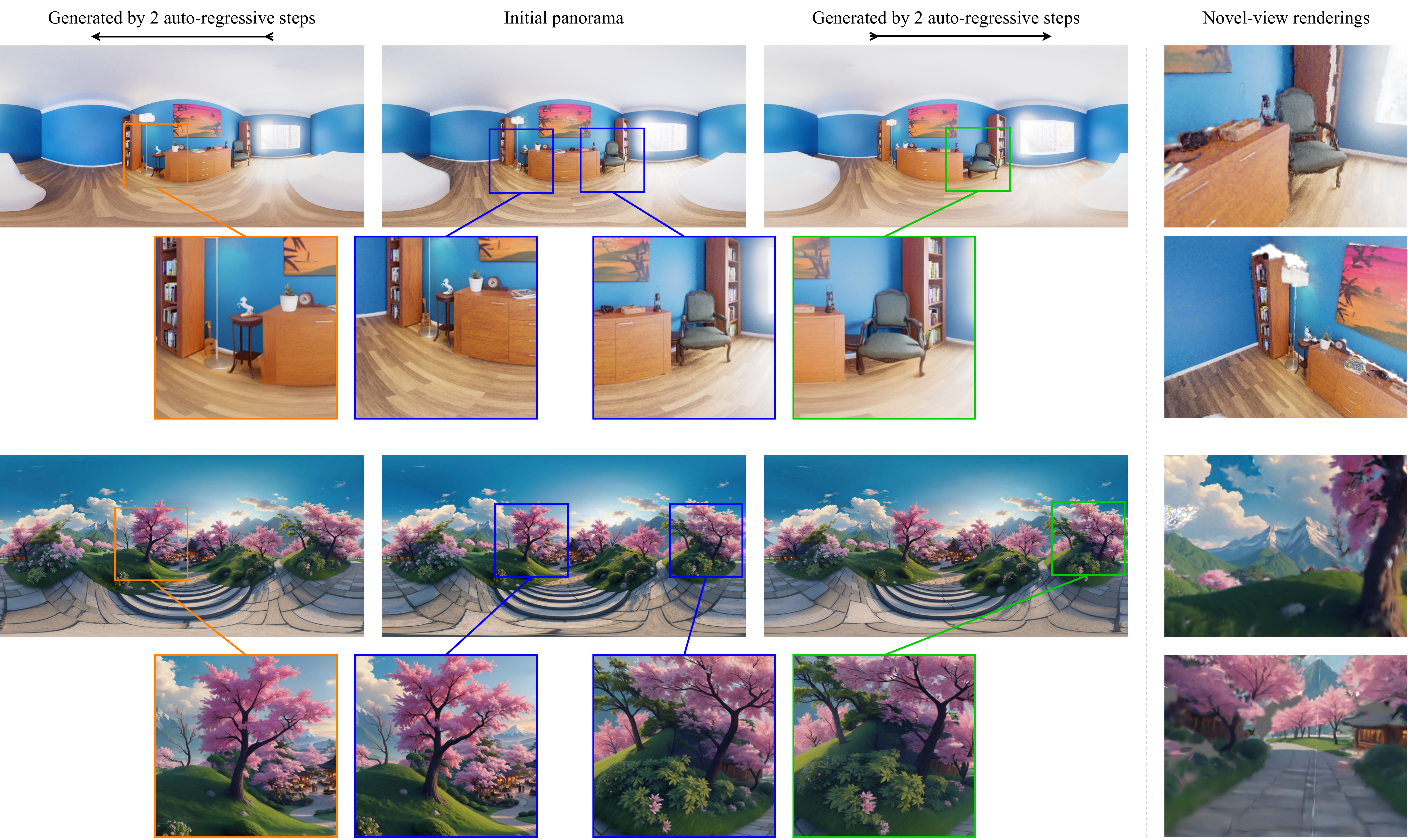}
\vspace{-.65cm}
\caption{\textbf{3D Scene Generation.} We provide visual example of generated novel-view panoramas on the left side. The details of the initial panorama are well preserved and previously unseen regions are filled in. On the right side we show novel-view renderings of the reconstructed scenes indicating the 3D consistency of the generated panoramas.}
\label{fig:qualitative_main}
\vspace{-.35cm}
\end{figure*}
Panoramic images are a powerful representation because they capture a significant part of a scene's context.
Unfortunately, existing panorama datasets are generally quite small and contain only a single image per scene.
To overcome this limitation, we thus develop a pipeline built on top of Infinigen~\cite{infinigen2023infinite, infinigen2024indoors} to generate a custom, synthetic dataset of multi-view panoramas.
Infinigen procedurally generates and populates indoor and outdoor scenes within the 3D rendering software Blender.
We adopt Infinigen to generate 3D scenes and render high-resolution multi-view 360 panoramas, as can be seen in \cref{fig:infinigen_data}.
In total, we collect around 230\,000 pairs of panoramas at a resolution of $4096 \times 2048$ across 5\,000 scenes to train our multi-view generation model.
Furthermore, we find that existing works in 3D scene generation do not have a unified system for quantitative evaluation and often rely on hand-crafted solutions.
Therefore, we also curate a small test set consisting of six photorealistic scenes from Blender and ten scenes generated with Infinigen.
For every scene, nine panoramas and their corresponding depth maps are rendered in an area of $[-1m, 1m]$ around the scene origin and serve as references for evaluating visual quality.
Both the training and testing dataset will be made publicly available to facilitate future research.

As introduced in \cref{sec:model}, our multi-view panorama generation model is built on a customized version of the popular LDM~\cite{rombach2022high} model.
It processes twelve individual cubemap faces at a resolution of $1024 \times 1024$, derived from two $4096 \times 2048$ input equirectangular panoramas.
We initialize our model with pretrained weights and finetune for 90,000 steps ($\sim 2.5$ days) at a batch size of 1 (= 12 cubefaces) sharded across 4 ViperFish TPUs with 64 TPUs in total, making for an effective batch size of 16. 
We empirically find that a step size of $d=0.25m$ provides a good trade-off between scene exploration and robustness.

\subsection{3D Scene Generation}

\begin{figure*}[!t]
\centering
\includegraphics[width=\linewidth]{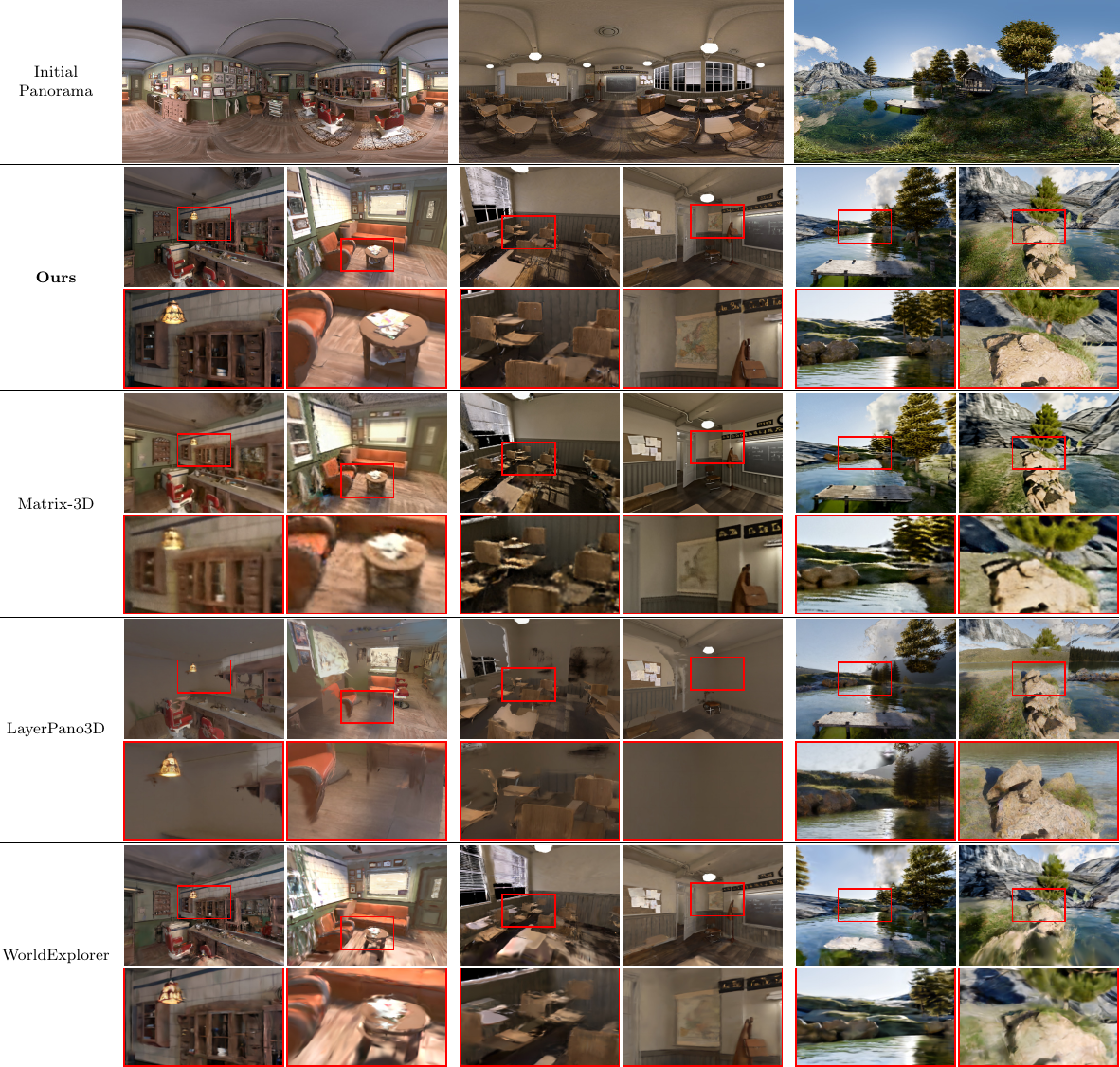}
\vspace{-.65cm}
\caption{\textbf{Comparison with Baselines.} Given a high quality input panorama, we observe that our approach achieves consistent scene generation while showing significantly more details and sharpness in the rendered novel view images in comparison to the baselines.}
\label{fig:baselines_qualitative}
\vspace{-.5cm}
\end{figure*}
\paragraph{Qualitative results.}
To provide an overview of the capabilities of our model, we show a range of scenes in \cref{fig:qualitative_main}, with their corresponding input, generated multi-view panoramas, and obtained Gaussian Splatting reconstructions.
As demonstrated, the generated novel-view panoramas retain all details from the initial panorama while correctly adjusting the geometry of the objects.
For example, the highlighted chair is correctly translated despite its complex geometric structure.
Furthermore, the previously occluded regions are well filled in, whilst respecting the overall context.
This shows that our multi-view panorama generation model not only learns a strong geometric understanding of the environment, but also retains its powerful generative inpainting capability.
The robust geometric reasoning capabilities are further underlined by the high-quality pointcloud, as produced by MapAnything.
Finally, the 3DGS renderings are also high quality, even for viewpoints that are far away from the initial panorama.

\paragraph{Comparisons with the state-of-the-art.}
We also perform qualitative and quantitative evaluations against several state-of-the-art baselines:
\textbf{LayerPano3D}~\cite{yang2025layerpano3d} iteratively removes and inpaints layers from an input panorama to build a multi-plane panorama image.
For detecting individual layers, they employ a depth and segmentation model.
\textbf{WorldExplorer}~\cite{schneider2025worldexplorer} starts with a panoramic image and generates videos along several predefined trajectories using a camera-conditioned video diffusion model.
To provide the scene context during generation, they sample previously generated frames and prepend them to the video.
\textbf{Matrix-3D}~\cite{yang2025matrix}, concurrent to us, fine-tunes a video diffusion model to generate panorama videos along a camera trajectory.
However, due to video generation models being significantly more expensive than image generation models, they are restricted to a maximum resolution of $1440 \times 720$ despite significant computational resources.

\begin{table}[t!]
\centering
\small
\newcolumntype{C}[1]{>{\centering\arraybackslash}p{#1}}
\begin{tabular}{l C{1.3cm} C{1.3cm} C{1.3cm}}
\toprule
\textbf{Infinigen Indoors} & PSNR$\uparrow$ & SSIM$\uparrow$ & LPIPS$\downarrow$ \\
\cmidrule(lr){2-4}
\textit{\hspace{1em}WorldExplorer} & 11.864 & 0.674 & 0.739 \\
\textit{\hspace{1em}LayerPano3D} & 18.305 & \underline{0.783} & 0.509 \\
\textit{\hspace{1em}Matrix-3D} & \underline{18.532} & 0.753 & \underline{0.502} \\
\textit{\hspace{1em}Ours} & \textbf{21.775} & \textbf{0.797} & \textbf{0.430} \\
\midrule
\textbf{Infinigen Outdoors} & PSNR$\uparrow$ & SSIM$\uparrow$ & LPIPS$\downarrow$ \\
\cmidrule(lr){2-4}
\textit{\hspace{1em}WorldExplorer} & 13.912 & 0.561 & 0.594 \\
\textit{\hspace{1em}LayerPano3D} & 17.364 & \underline{0.590} & 0.537 \\
\textit{\hspace{1em}Matrix-3D} & \underline{17.970} & 0.581 & \underline{0.529} \\
\textit{\hspace{1em}Ours} & \textbf{20.507} & \textbf{0.646} & \textbf{0.384} \\
\midrule
\textbf{Blender Scenes} & PSNR$\uparrow$ & SSIM$\uparrow$ & LPIPS$\downarrow$ \\
\cmidrule(lr){2-4}
\textit{\hspace{1em}WorldExplorer} & 13.659 & 0.637 & 0.611 \\
\textit{\hspace{1em}LayerPano3D} & \underline{18.124} & \underline{0.692} & \underline{0.463} \\
\textit{\hspace{1em}Matrix-3D} & 17.898 & 0.660 & 0.515 \\
\textit{\hspace{1em}Ours} & \textbf{21.995} & \textbf{0.762} & \textbf{0.342} \\
\midrule
\textbf{Average} & PSNR$\uparrow$ & SSIM$\uparrow$ & LPIPS$\downarrow$ \\
\cmidrule(lr){2-4}
\textit{\hspace{1em}WorldExplorer} & 13.145 & 0.624 & 0.648 \\
\textit{\hspace{1em}LayerPano3D} & 17.931 & \underline{0.688} & \underline{0.503} \\
\textit{\hspace{1em}Matrix-3D} & \underline{18.133} & 0.665 & 0.515 \\
\textit{\hspace{1em}Ours} & \textbf{21.426} & \textbf{0.735} & \textbf{0.385} \\
\bottomrule
\end{tabular}
\vspace{-.2cm}
\caption{\textbf{Quantitative Evaluation.} We compare our approach to the baseline methods for three different datasets on common image metrics. We observe that our approach yields significant improvements on all metrics and datasets.}
\label{tab:main_nvs}
\vspace{-.5cm}
\end{table}

We provide every method with the same initial panorama and a text prompt whenever applicable.
After generation, we align the scenes to the ground-truth scale by comparing rendered and ground-truth depth maps.
In \cref{fig:baselines_qualitative}, we show a qualitative comparison against the baselines.
First, LayerPano3D provides fairly sharp renderings, however, quality degrades in occluded areas as the automatic layering of the scene oftentimes struggles, leading to artifacts and inconsistencies with the initial panorama.
For example, the mountain background in the barbershop is replaced with a different door and the lowest layer (brown mist) also blends into the scene.
The scenes generated by WorldExplorer are reasonable, but contain a fairly low level of detail.
Especially when moving further away from the initial panorama, one experiences severe Gaussian defects.
We attribute this to the fact that the generated videos experience drift over time, as well as local color inconsistencies.
Finally, Matrix-3D generates very consistent and robust results. 
Nonetheless, the resulting scenes lack details and often appear blurry. 
We hypothesize that this is a result of the (for panoramic images) very low resolution of $1440 \times 720$.
In contrast to them, our generated scenes are generally sharp and remain consistent even when moving further away.
These findings are also consistent with our quantitative evaluation, which we present in \cref{tab:main_nvs}.
Following common practice, we report the standard NVS metrics PSNR, SSIM, and LPIPS, comparing renderings from the generated Gaussians with the ground-truth views.
As can be easily observed, our method clearly outperforms all baselines across all datasets and metrics.
For example, we outperform the state of the art by at least $3.3$ dB on average in PSNR.
Similarly, for SSIM and LPIPS, Stepper achieves strong results of $0.735$ and $0.385$, clearly exceeding the second-best method, LayerPano3D, with $0.688$ and $0.503$, respectively.

\subsection{Ablation Studies}

\begin{figure}[t!]
\centering
\includegraphics[width=.9\linewidth]{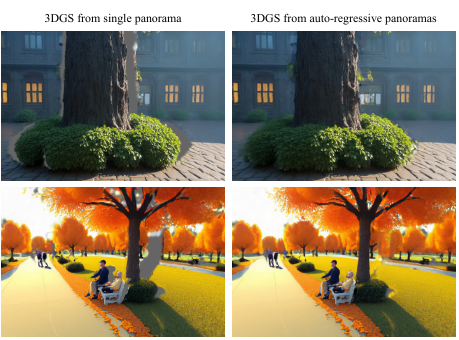}
\vspace{-.3cm}
\caption{\textbf{Single vs multiple panoramas to 3DGS.} The multi panorama input to the 3DGS reconstruction consistently fills in the unobserved regions in the initial panorama without sacrificing the quality of the input panorama.}
\label{fig:inpainting}
\vspace{-.3cm}
\end{figure}
\paragraph{Auto-regressive expansion.}
One of our main contributions is an auto-regressive scene expansion scheme via novel-view panorama synthesis, which enables an immersive experience.
To further underline its necessity, we compare our final 3DGS scenes against scenes obtained from only the initial panorama and the respective scene geometry.
As can be seen in \cref{fig:inpainting}, the scene obtained from only a single panorama has significantly more gaps.
In contrast, our pipeline can retain all details from the initial scene whilst significantly improving completeness.

\begin{figure}[t!]
\centering
\includegraphics[width=\linewidth]{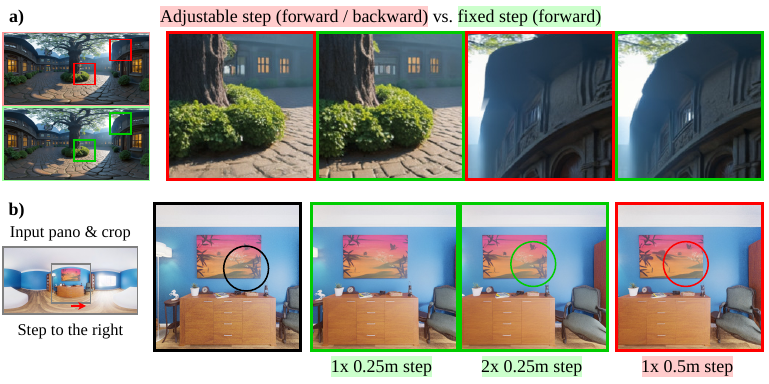}
\vspace{-.75cm}
\caption{\textbf{Effect of step size.} Novel panos.\ generated by a model with \textbf{a)} adjustable step direction, \textbf{b)} a larger step size $d=0.5m$.}
\label{fig:step_size}
\vspace{-.3cm}
\end{figure}
\paragraph{Analysis of step size.}
To test the effect of the fixed step size, we train two further model variants: 
\textbf{a)} A model for both forward and backward stepping, with the direction provided via a conditioning signal, and \textbf{b)} a model with a step size of $d=0.5m$ (\textit{vs} the default $d=0.25m$).
As can be seen in \cref{fig:step_size}, the adjustable-step model sometimes produces wrong geometry and poor textures with artifacts, while our fixed-step model does not suffer from these issues.
We attribute this to the learning task being easier when using a fixed step and thus selected this architecture.
The $d=0.5m$ model still generates high-quality novel panoramas, but is slightly worse at retaining details.
The default $d=0.25m$ model offers a good trade-off between step granularity, panorama quality, and exploration.

\section{Conclusion}
\label{sec:conclusion}

We introduced \emph{Stepper}, a framework for text-driven immersive 3D scene generation that addresses the trade-off between visual fidelity and explorability. 
By combining a multi-view 360$^\circ$ diffusion model with a feed-forward reconstruction pipeline and a large-scale dataset of $230 000$ multi-view panoramas, Stepper generates high-quality, large-baseline explorable scenes without the context drift of prior methods. 
Our experiments demonstrate significant improvements over recent baselines, achieving an average PSNR improvement of $3.3$dB, thereby establishing a new standard for immersive 3D scene synthesis.

\newpage
{
    \small
    \bibliographystyle{ieeenat_fullname}
    \bibliography{main}
}

\clearpage
\setcounter{page}{1}
\maketitlesupplementary
\appendix

\section{Overview}
\label{sec:overview}

This supplementary material will provide \textbf{additional results} in \cref{sec:supp_add_results}, including 
several video results, 
a user study, and further ablation studies.
We also provide more details regarding the \textbf{technical implementation} in \cref{sec:supp_technical_details}, the \textbf{datasets} in \cref{sec:supp_datasets}, and the \textbf{evaluation protocol} in \cref{sec:supp_evaluation_details}.

\section{Additional Results}
\label{sec:supp_add_results}

\subsection{Video Results}
To highlight the differences between our method and the baselines, we generate the same scene with every method and render trajectories from it. 
These renderings are provided on our project page at \href{https://fwmb.github.io/stepper}{fwmb.github.io/stepper}.

\subsection{User Study}
\begin{figure}[t!]
\centering
\includegraphics[width=\linewidth]{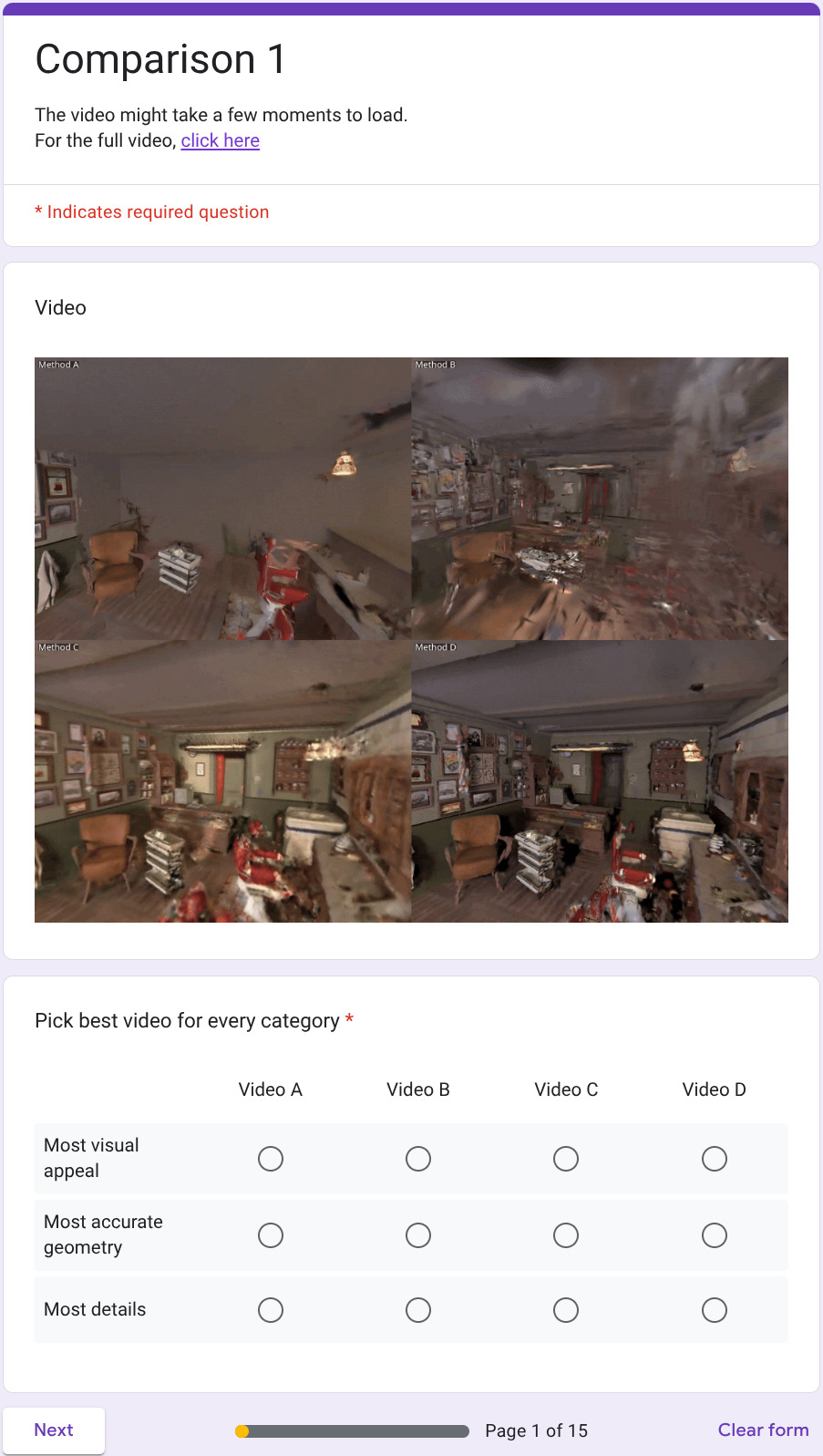}
\caption{\textbf{User study form.} Participants fill out this randomized form and cast a single vote per category.}
\label{fig:user_study_form}
\end{figure}

Additionally to our quantitative evaluation, we also perform a user-study.
Here, $n=10$ participants, who were not familiar with the project, are shown five scenes with three different camera trajectories each (15 samples).
For every sample, we randomly arrange our method and the three baselines.
The participants then pick the best video for the three categories 1) \textit{most visually appealing}, 2) \textit{most accurate geometry}, and 3) \textit{most details}.
A screenshot of the study form can be seen in \cref{fig:user_study_form}.

We collected 150 votes per category and report the results in \cref{tab:user_study}, which support our qualitative and quantitative findings from the main paper.
Overall, our method clearly produces the most appealing scenes. 
Matrix-3D is a strong competitor, however, it falls short especially on the \textit{most details} category.
This aligns with our insight that panorama video models cannot operate on resolutions high enough to obtain sharp results.
The other two baselines LayerPano3D and WorldExplorer are far behind.
We observe that for LayerPano3D, the layering often fails to properly disentagle the context of the scene, leading to strong blending artifacts.
WorldExplorer generally struggles to retain scene consistency when moving further away from the origin, resulting in distorted images and Gaussian artifacts.
In general, the user study confirms the results from our qualitative and quantitative evaluation in the main paper and further proves the efficiency of our new method.

\subsection{Ablations}
\begin{figure*}[t!]
\centering
\includegraphics[width=\linewidth]{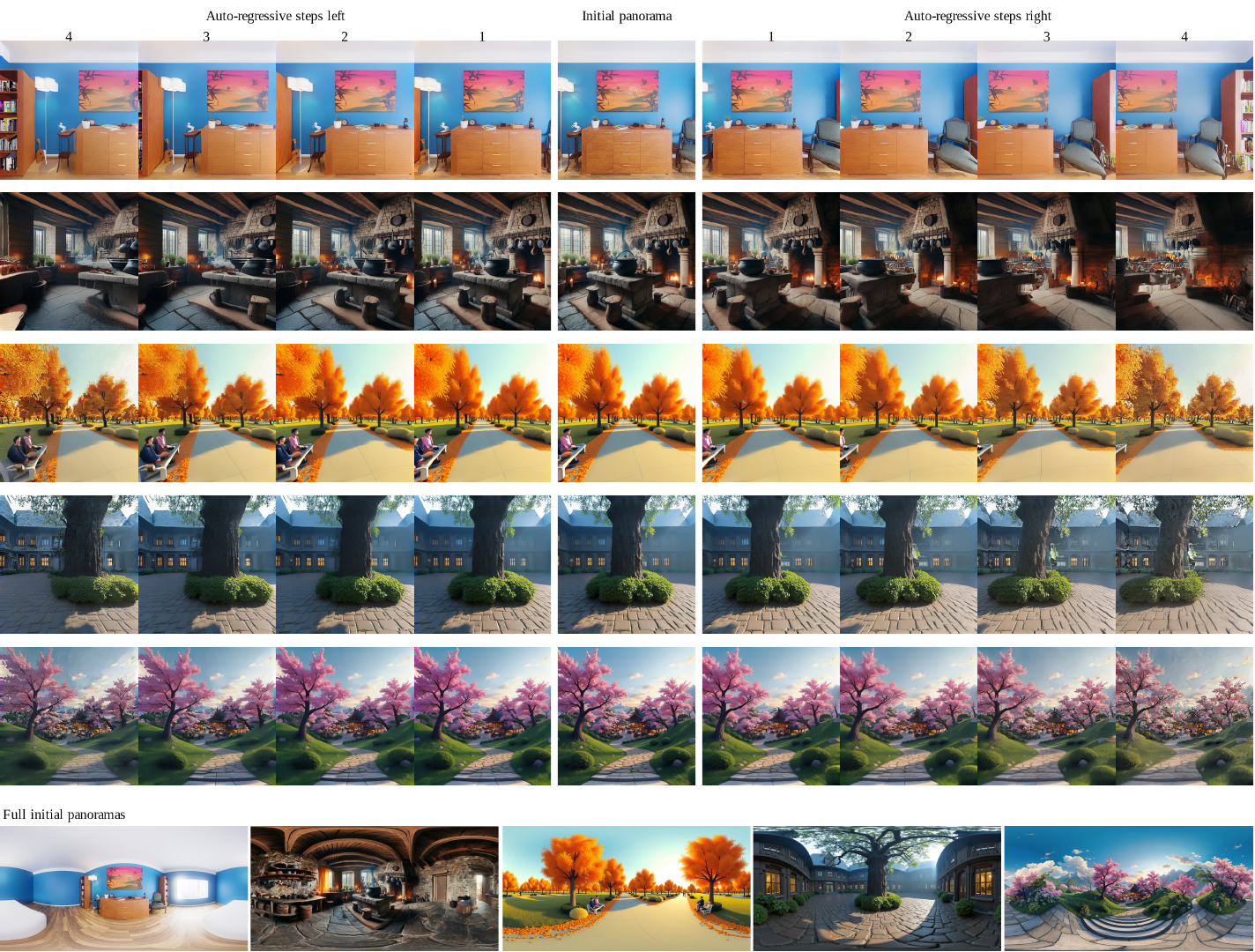}
\caption{\textbf{Auto-regressive steps.} We perform a number of auto-regressive steps to generate novel-view panoramas from the initial panorama to the left and right for several examples. From the generated panoramas, we visualize the forward-facing view to highlight the generated details. A single step corresponds to a step length of $0.25m$.}
\label{fig:supp_unrolling}
\end{figure*}

\paragraph{Auto-regressive stepping.}
To test the robustness of our model, we auto-regressively apply our model for a large number of steps and visualize the results in \cref{fig:supp_unrolling}.
As can be seen, our model is able to predict meaningful novel views even for further away steps.
However, after the third or fourth step, the visual quality starts to degrade slightly and artifacts can appear.
For our main experiments, we opt to perform $n=2$ steps in every direction (front, right, back, left), as this ensures maximum quality of our generated novel panorama views, while covering a large area.
We believe that enabling the model to do even more steps would be a promising direction for future work.
\begin{table}[]

\centering
\footnotesize

\definecolor{colorOurs}{HTML}{e76f51}
\definecolor{colorMatrix}{HTML}{f4a261}    %
\definecolor{colorLayer}{HTML}{e9c46a}     %
\definecolor{colorWorld}{HTML}{2a9d8f}     %

\centering
\footnotesize

\hspace{-22pt}
\begin{tikzpicture}
    \begin{axis}[
        ybar,
        width=\linewidth, height=5cm,      %
        bar width=12pt,              %
        enlarge x limits=0.2,        %
        legend style={at={(0.5,-0.2)},
          anchor=north,legend columns=-1}, 
        ylabel={Votes},
        ylabel style={yshift=-15pt},
        symbolic x coords={Visual Appeal, Geometry, Details},
        xtick=data,
        nodes near coords,
        nodes near coords align={vertical},
        ymin=0,
        ymajorgrids=true,
        grid style=dashed,
    ]
        \addplot[fill=colorOurs, draw=none] coordinates 
            {(Visual Appeal,88) (Geometry,78) (Details,98)};
        
        \addplot[fill=colorMatrix, draw=none] coordinates 
            {(Visual Appeal,59) (Geometry,62) (Details,47)};
        
        \addplot[fill=colorLayer, draw=none] coordinates 
            {(Visual Appeal,1) (Geometry,9) (Details,2)};
            
        \addplot[fill=colorWorld, draw=none] coordinates 
            {(Visual Appeal,2) (Geometry,1) (Details,3)};

        \legend{Ours, Matrix3D, LayerPano3D, WorldExplorer}
    \end{axis}
\end{tikzpicture}

\vspace{5pt}

\begin{tabular}{lccc}
\toprule
\textbf{Votes}         & \textbf{Visual Appeal} & \textbf{Geometry} & \textbf{Details} \\
\midrule
\textit{WorldExplorer} & 2                      & 1                 & 3                \\
\textit{LayerPano3D}   & 1                      & 9                 & 2                \\
\textit{Matrix3D}      & 59                     & 62                & 47               \\
\textit{Ours}          & \textbf{88}                     & \textbf{78}                & \textbf{98}               \\
\bottomrule
\end{tabular}
\caption{\textbf{User study results.} $n=10$ participants cast votes on the most \textit{visually appealing}, \textit{most accurate geometry} and \textit{most details} categories for 15 video comparisons, yielding 150 votes per category in total.}
\label{tab:user_study}
\end{table}

\begin{figure}[t!]
\centering
\includegraphics[width=\linewidth]{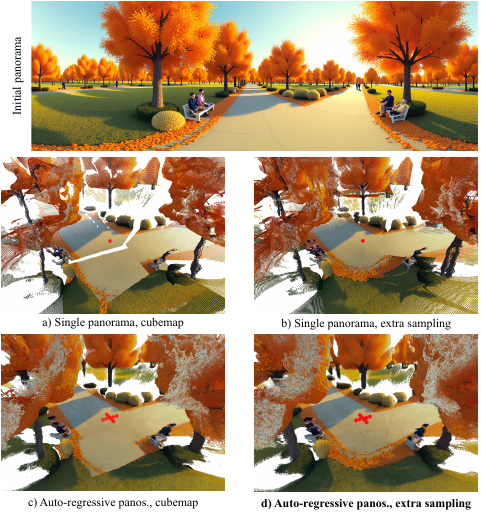}
\caption{\textbf{Impact of Multi-View Panoramas} We depict the effect of using various options of panorama input for MapAnything and found that the auto-regressive expansion and sampling yield the most complete and accurate outputs.}
\label{fig:mapanything_ablation}
\end{figure}

\paragraph{Effectiveness of MapAnything.}
The three baselines methods that we compare against rely on a monocular depth predictor to obtain initial scene geometry from the initial panorama.
In contrast, we first generate several novel panorama views using our model, and then apply MapAnything as our reconstruction model.
To test whether the usage of multiple panorama views has an effect on the reconstruction capabilities of MapAnything, we test it in different configurations, as can be seen in \cref{fig:mapanything_ablation}.
When applying MapAnything only on a single cubemap (a), MapAnything fails to properly align the pointmaps of the different views. 
Sampling more views from the initial panorama (b) helps, but there are still several problematic parts (e.g. bushes in the back).
Multiple cubemaps from panoramas from auto-regressive stepping (c) provide a strong stereo signal and result in very consistent and accurate geometry reconstruction. 
However, we find that MapAnything struggles with textureless regions with cameras facing downwards.
This is fixed by again sampling more views from the different generated panoramas (d).

\paragraph{Runtime analysis.}
While our pipeline is not optimized for inference speed yet, it is significantly faster than the baselines, as shown in details in \cref{tab:timing}.
There are several possible optimizations: 
\textit{1)} The 3DGS optimization can be replaced by a feed-forward model like AnySplat.
\textit{2)} The panorama generation model currently does 50 diffusion steps. 
Using improved solvers and distillation, this can be reduced significantly.
\textit{3)} Scene expansion could be done \textit{on-demand}, only generating novel panoramas in areas where needed.
\begin{table}[]
\centering
\footnotesize
\begin{tabular}{lc|lc}
\toprule
\multicolumn{2}{c}{\textit{Our stages (single step = 30s)}} & \multicolumn{2}{c}{\textit{Ours vs Baselines}} \\
\cmidrule(lr){1-2}\cmidrule(lr){3-4}
1. Gen. $n=8$ novel panos       & $n\times$30s       & WorldExplorer             & 7h             \\
2. MapAnything       & 20s      & Matrix-3D            & 40min          \\
3. Filtering         & 2min       & LayerPano3D          & 32min          \\
4. 3DGS optimization              & 8min       & \textbf{Ours}                 & \textbf{15min} \\
\bottomrule
\end{tabular}
\caption{\textbf{Inference time - Panorama to 3D scene.}}
\label{tab:timing}
\end{table}

\section{Technical Details}
\label{sec:supp_technical_details}

For all of our experiments, we use the same multi-view panorama generation model.
It was obtained via finetuning a pre-trained CubeDiff model, which itself is based on a variant of the popular Latent Diffusion Model architecture.
The finetuning uses the default diffusion loss for 90000 steps.
The learning rate is set to $8e^{-5}$, with a linear ramp-up from 0 during the first 10000 steps.
During finetuning and inference of the model, we remove all text-conditioning.
Generating a single novel-view panorama with 50 diffusion steps takes around 30 seconds.
For MapAnything, we rely on the commercial checkpoint under Apache license.
To improve the visual quality of the renderings, we add a skybox texture to the 3DGS scene where MapAnything predicts infinite distance.

\section{Datasets}
\label{sec:supp_datasets}

\begin{figure*}
    \centering
    \includegraphics[width=\linewidth]{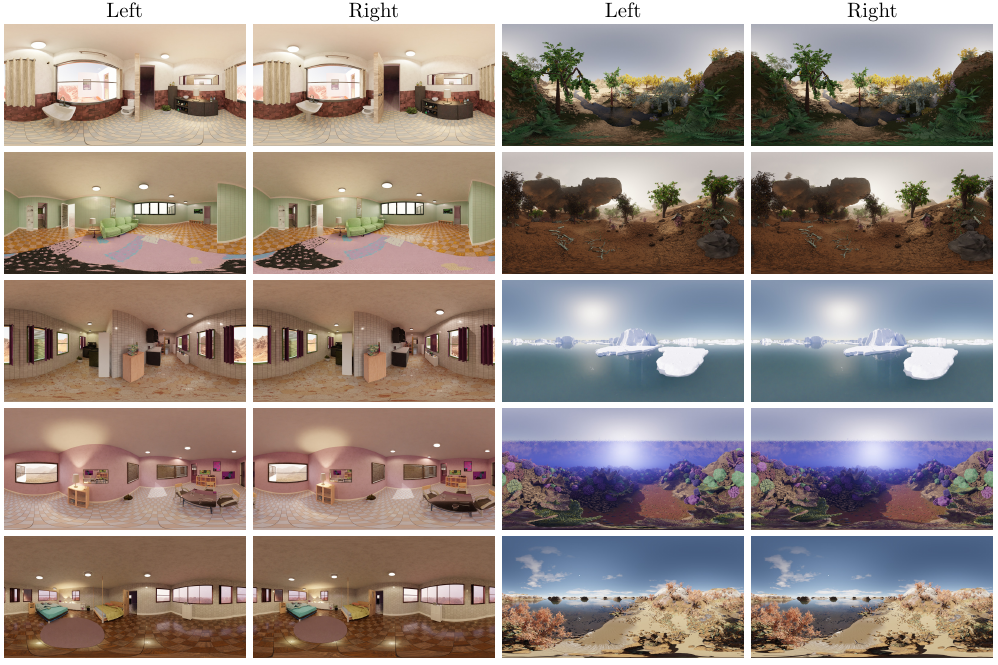}
\caption{\textbf{Multi-view panoramas pairs generated with Infinigen.}}
\label{fig:supp_data}
\end{figure*}

\paragraph{Training data.}
As described in the main paper, we adapt the Infinigen framework to procedurally generate synthetic multi-view panorama data, as can be seen in \cref{fig:supp_data}.

Indoor scenes are rooms of different categories (livingroom, bathroom, bedroom, kitchen, \etc) and require about 1 hour to generate on a standard CPU machine.
Outdoor scenes are also sampled from different biomes (river, mountains, forest, cave, arctic, \etc).
However, due to them requiring more, complex assets like trees and bushes, their generation takes up to 3 hours on CPU.
In total, we generate 4730 indoor and 3252 outdoor scenes

During the generation process of a scene, a number of positions are sampled for panoramic cameras. 
For an indoor scene, the pipeline places up to $n = 20$ cameras at random locations in the room.
To reduce the numbers of assets that have to be generated to cover the bigger outdoor scenes, we only place $n = 4$ cameras.
Sometimes, the placement algorithm can only find a smaller number of cameras.
For every placed camera, we render its view and the views $d=0.25m$ to the right and to the left.
The two pairs $[[\text{left}, \text{center}], [\text{center}, \text{right}]]$, as well as the mirrored pair $[[\text{center}, \text{left}],[\text{right}, \text{center}]]$, where we rotate the panorama by $180^{\circ}$, are rendered.
Therefore, every placed camera yields $m = 4$ pairs of left-right panoramas with our desired baseline.
In total, we generate 187156 indoor and 49372 outdoor panorama pairs.
Due to the high resolution of $4096 \times 2048$ and raytracing, the rendering process takes around 70 seconds per panorama on an A100 GPU.

\begin{figure*}
    \centering
    \includegraphics[width=\linewidth]{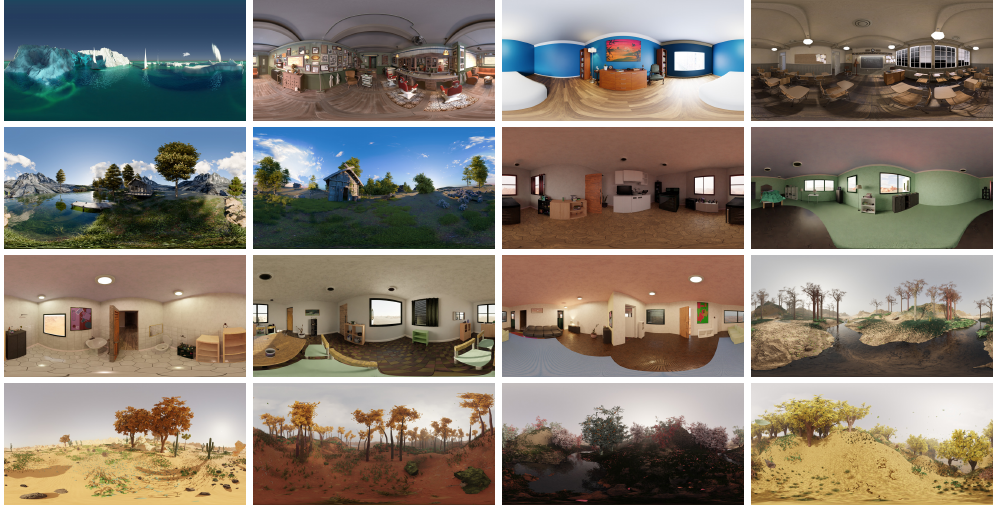}

\caption{\textbf{Input panoramas used for testing.}}
\label{fig:supp_data_test}
\end{figure*}

\paragraph{Testing data.}
We curate a test set of 16 synthetic scenes, for which we render an initial panorama and eight novel-view panoramas from different locations at a resolution of $4096\times 2048$.
For each panorama, we also compute the corresponding depth map.
The scenes are obtained using Blender (6), Infinigen indoors (5) and Infinigen outdoors (5) and are visualized in \cref{fig:supp_data_test}.
For the Blender scenes, we adapt the following sources, which are all under Creative Commons licenses:
Arctic ships\footnote{\href{https://download.blender.org/archive/gallery/blender-splash-screens/blender-3-2/}{https://download.blender.org/archive/gallery/blender-splash-screens/blender-3-2/}}, 
Barbershop\footnote{\href{https://svn.blender.org/svnroot/bf-blender/trunk/lib/benchmarks/cycles/barbershop_interior/}{https://svn.blender.org/svnroot/bf-blender/trunk/lib/benchmarks/\\cycles/barbershop\_interior/}}, 
Classroom\footnote{\href{https://www.blender.org/download/demo-files/}{https://www.blender.org/download/demo-files/}}, 
Mountain cabin\footnote{\href{https://www.blenderkit.com/get-blenderkit/28cb035a-d333-4e7f-a2b5-543629cdd982/}{https://www.blenderkit.com/get-blenderkit/28cb035a-d333-4e7f-a2b5-543629cdd982/}}, 
Valley\footnote{\href{https://www.blenderkit.com/get-blenderkit/28cb035a-d333-4e7f-a2b5-543629cdd982/}{https://www.blenderkit.com/get-blenderkit/28cb035a-d333-4e7f-a2b5-543629cdd982/}}, 
Blue room\footnote{\href{https://blog.polyhaven.com/blue-wall-scene-file/}{https://blog.polyhaven.com/blue-wall-scene-file/}}.

\paragraph{Qualitative examples.}
We rely on a number of qualitative examples from other papers\footnote{\href{https://ys-imtech.github.io/projects/LayerPano3D/}{https://ys-imtech.github.io/projects/LayerPano3D/}}\footnote{\href{https://katjaschwarz.github.io/worlds/}{https://katjaschwarz.github.io/worlds/}} to demonstrate the generalization capabilities of our method.

\section{Evaluation Details}
\label{sec:supp_evaluation_details}

\paragraph{Setup.}
For every test scene, we generate a 3DGS scene with every baseline. 
As a first step, we transform every scene such that its scene origin and rotation line up with the ground-truth scene.
This can be achieved using the camera configurations from the individual reconstructions.
Then, we align the scale of every scene with the scale of the ground-truth scale.
To this end, we render a depth map in all four horizontal directions with a 90 degree field-of-view and a resolution of $1024\times1024$ for the initial panorama (at the origin of the scene).
After filtering out invalid depths, the scene scale is determined via the median of all scales between the rendered and ground-truth depths.
Our scenes tend to contain around $2.5 \times 10^6$ Gaussians, Matrix-3D's scenes around $10 \times 10^6$ Gaussians, LayerPano3D's scenes around $3 \times 10^6$ Gaussians, and WorldExplorer's scenes around $0.4 \times 10^6$ Gaussians.

\paragraph{Rendering.}
Using the alignment and scaling from above, we render the six cubefaces for each of the 9 panorama (initial + 8 novel-view panoramas) with a field of view of 90 degrees and a resolution of $1024 \times 1024$ once from the 3DGS scene and once from the ground-truth panoramas using resampling.
The 3DGS rendering is performed using the open-source \texttt{gsplat} library.
For every rendered cubeface (54 in total), we compute the NVS metrics SSIM, LPIPS, and PSNR and report the average.


\end{document}